\documentclass[letterpaper, 10 pt, journal, twoside]{IEEEtran} 
\IEEEoverridecommandlockouts 

\usepackage{verbatim}
\usepackage[pdftex]{graphicx}
\usepackage{epsfig,subfigure}
\usepackage{amsmath,amssymb,amsfonts,bm,amscd} 
\usepackage{mathrsfs}
\usepackage{setspace}
\usepackage{fancyhdr}
\usepackage{algorithm} 
\usepackage{psfrag}
\usepackage{cite}
\usepackage{makecell}
\usepackage{url}
\usepackage{float}
\usepackage{soul}
\usepackage{color}

\usepackage{color, xcolor}

\begin{document}

\title{Learning-based Predictive Path Following Control for Nonlinear Systems Under Uncertain Disturbances}

\author{Rui Yang$^{1}$, Lei Zheng$^{2}$, Jiesen Pan$^{1}$, Hui Cheng$^{1}$
	\thanks{Manuscript received: October, 15, 2020; Revised January, 12, 2021; Accepted February, 4, 2021.}
	\thanks{This paper was recommended for publication by Editor Dana Kulic upon evaluation of the Associate Editor and Reviewers' comments.}
	\thanks{$^{1}$R. Yang, J. Pan, H. Cheng are with the School of Computer Science and Engineering, Sun Yat-Sen University,
		Guangzhou, China. Corresponding author: H. Cheng
		{\tt\footnotesize chengh9@mail.sysu.edu.cn}.}
	\thanks{$^{2} $L. Zheng  is with the School of Electronics and Information Technology, Sun Yat-sen University, Guangzhou, China.}
	\thanks{Digital Object Identifier (DOI) 10.1109/LRA.2021.3062805.}
}%

\markboth{IEEE ROBOTICS AND AUTOMATION LETTERS, VOL. 6, NO. 2, APRIL 2021
}
{Yang \MakeLowercase{\textit{et al.}}: Learning-based Predictive Path Following Control for Nonlinear Systems Under Uncertain Disturbances}

	\maketitle
	\begin{abstract}
		Accurate path following is challenging for autonomous robots operating in uncertain environments. Adaptive and predictive control strategies are crucial for a nonlinear robotic system to achieve high-performance path following control. In this paper, we propose a novel learning-based predictive control scheme that couples a high-level model predictive path following controller (MPFC) with a low-level learning-based feedback linearization controller (LB-FBLC) for nonlinear systems under uncertain disturbances. The low-level LB-FBLC utilizes Gaussian Processes to learn the uncertain environmental disturbances online and tracks the reference state accurately with a probabilistic stability guarantee. Meanwhile, the high-level MPFC exploits the linearized system model augmented with a virtual linear path dynamics model to optimize the evolution of path reference targets, and provides the reference states and controls for the low-level LB-FBLC. Simulation results illustrate the effectiveness of the proposed control strategy on a quadrotor path following task under unknown wind disturbances.
	\end{abstract}
	\section{Introduction}
	\label{sec:introd}
	\IEEEPARstart{I}{ncreasingly} wide applications of autonomous mobile robots, such as in package delivery, electrical lines supervision and industrial inspection, require the controller to handle unpredictable and potentially adverse outdoor conditions and maintain high trajectory control performance. In particular, robots should be accurately steered along a predefined path in the presence of uncertain environmental disturbances. These non-negligible uncertainties are usually hard to model, such as the aerodynamic effects of flying vehicles. This makes it difficult for high-performance trajectory control using model-based controllers. Besides, it is also not realistic to tune the controller parameters manually for each specific operating condition. Desired trajectory controllers therefore should exhibit high control accuracy, adapt online to the uncertain environment, be robust to the uncertain disturbances, and take input constraints into account.
 
	The trajectory control problem, defined as steering a robot to follow a predefined path, can be solved using trajectory tracking or path following~\cite{matschek2019nonlinear}. Compared with the trajectory tracking approach, where a timed parameterized reference is tracked, the path following approach removes the time dependence and regards it as a degree of freedom in the controller design. It steers the robot along a geometric path while prioritizing closeness to the desired reference with an attempt to satisfy a dynamic specification, such as a speed assignment along the path. This feature brings in some advantages to control performance and exhibits robustness to disturbances~\cite{matschek2019nonlinear}. For example, if the robot loses track of the path under irresistible environmental disturbances, it will slow down to come back to the path rather than attempt to align with the time-dependent reference. 
	\begin{figure}[t]
		\begin{center}
			\includegraphics[scale=0.22]{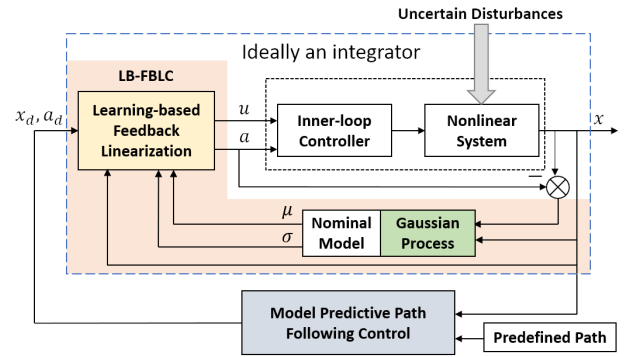} 
		\end{center} \vspace{-3mm}
		\caption{Architecture diagram of the proposed strategy for nonlinear system path following under uncertain disturbances. The GPs are used to learn the disturbances online with uncertainty bounds. With the estimation and the uncertainty bound, the low-level LB-FBLC forces the nonlinear system to behave like an integrator and tracks the reference state $x_d=[x_{1d},x_{2d}]^\mathrm{T}$. The linear integrator is used as a predictive model in the high-level MPFC, which optimizes the reference target evolution along the path and provides reference states $x_d$ and reference controls $a_d$ for the LB-FBLC.} 
		\vspace{-7mm}
		\label{fig:structure}
	\end{figure}
	
	Typical path following approaches usually do this by separating the control scheme into an outer-loop guidance law for the generation of the reference motions along the path and then an inner-loop controller to track those motions~\cite{rubi2019survey}. Among these methods, geometric methods, such as Carrot-chasing~\cite{micaelli1993trajectory} and nonlinear guidance law~\cite{park2007performance}, are widely used in nonlinear mobile robots path following tasks~\cite{rubi2019survey}. The reference target is chosen based on geometric methods and can not be optimized according to the path following error. Model predictive path following control (MPFC) shapes the path following problem into an optimal control problem of a model predictive control (MPC) framework~\cite{faulwasser2015nonlinear}. It uses the nonlinear system dynamics augmented with a virtual path dynamics model to optimize the evolution of the reference target and reference controls. It can significantly improve the control performance by looking ahead to account for changes in the path~\cite{faulwasser2016implementation}, but the resulting MPC for the nonlinear system is subject to nonlinear models. The model error due to the uncertain disturbances may accumulate in multi-step prediction in MPC.

	For feedback linearizable nonlinear systems~\cite{khalil2002nonlinear}, feedback or feedforward linearization techniques can be used to convert the nonlinear dynamics into a linear integrator model, which can be used as the predictive model in MPC. Exploiting differential flatness property,~\cite{greeff2018flatness} combines the MPC with feedforward linearization control to tackle the trajectory tracking problem for a quadrotor. The MPFC technique combined with non-linear dynamic inversion acceleration controller is designed for multirotor path following in \cite{niermeyer2016geometric}. However, the performance of these model-based methods is still limited by the discrepancy between the nominal model and the actual system in the presence of unknown environmental disturbances. 

	To solve this problem, the uncertain disturbances should be estimated and compensated in controller design. In~\cite{8968471}, external wind gusts and ground effects of the quadrotor are estimated using the Kalman filter and compensated in the predictive model of MPC. Based on \cite{niermeyer2016geometric}, model reference adaptive control is introduced to the acceleration controller in \cite{akkinapalli2016adaptive}, but the adaptive scheme is limited to the rotational dynamics for a quadrotor.

	One alternative solution is to use data-driven machine learning methods to learn environmental disturbances and use the learned model to design the controller. In \cite{shi2019neural}, a deep neural network with spectral normalization is designed to learn the aerodynamic effects of a quadrotor in a feedback linearization controller to improve the quadrotor position control. Since only limited data can be obtained to approximate the uncertain disturbances, Gaussian Processes (GPs)~\cite{rasmussen2003gaussian} can be utilized to estimate the uncertain disturbances and capture both the epistemic uncertainty of the estimation due to the lack of data and the aleatoric uncertainty inherent in the environment~\cite{urbina2011quantification}. In~\cite{helwa2019provably}\cite{beckers2019stable}, the estimation using GPs and the corresponding confidence bounds are used to design a robust controller for Euler-Lagrange systems. While these proposed control strategies provide tracking control stability, control constraints are not considered in the controller. In~\cite{wang2018safe}, GPs are used in a feedback linearization controller to compensate the effect of wind disturbances. The controller reduces the trajectory tracking error for the quadrotor but without providing a control stability guarantee.
	
	To eliminate the effect of environmental disturbances and maintain the predictive capability of MPC, a nonlinear MPC is designed using GPs to update the model error of the nominal model for multi-step predictions\cite{mehndiratta2020gaussian}. In~\cite{pereida2018adaptive}, an $\mathcal{L}_1$ adaptive controller is designed to force the system to behave in a predefined linear model. A higher-level linear MPC uses this reference model to calculate the optimal reference input for the $\mathcal{L}_1$ controller for trajectory tracking problem.

	The limitations of the mentioned path following approaches and the advantages of learning-based control techniques show an urgent need for designing an adaptive and predictive path following control strategy that achieves accurate path following for nonlinear robotic systems under uncertain disturbances.

	In this paper, we propose a novel learning-based predictive path following control strategy, as illustrated in Fig. \ref{fig:structure}, for nonlinear systems to accurately follow the predefined paths under environmental disturbances. The control strategy couples an MPFC with an LB-FBLC which uses GPs to learn the environmental disturbances model online. The LB-FBLC forces the nonlinear system under disturbances to behave like an integrator and tracks the reference states. The high-level MPFC with the linear integrator model optimizes the evolution of the reference targets, and provides reference states and controls for the low-level LB-FBLC.

	The main contributions of the paper include:
	\begin{itemize}
		\item A novel online LB-FBLC combing first-principle and data-driven design methods is derived to enable accurate tracking control under uncertain environmental disturbances. The theoretical analysis of probabilistic stability is provided.
		\item A novel adaptive predictive path following control scheme coupling a high-level MPFC with the low-level LB-FBLC is designed to tackle path-following problems for the nonlinear system under unknown environmental disturbances. 
		%
		\item The effectiveness of the proposed control scheme is validated on the quadrotor path following task under uncertain wind disturbances via simulation.
	\end{itemize}

	The remainder of this paper is organized as follows. Problem statement and necessary preliminary are presented in Section~\ref{sec:problem}. The proposed methodology is introduced in Section~\ref{sec:alg}. In Section~\ref{sec:sim}, the effectiveness of the proposed method is validated via simulations on a quadrotor. The conclusion is reached in Section~\ref{sec:con}.

	\section{Problem Statement and Preliminary}
	\label{sec:problem}
	\subsection{Problem Statement}
	\label{sec:problem statement}
	Consider a nonlinear system with dynamics:	
	\begin{equation}
		\label{eq:sys_man}
		{\dot{x}}_1=x_2,\ {\dot{x}}_2=f(x)+G(x)u, 
	\end{equation}
	where the state $x=[x_1,x_2]^\mathrm{T} \in X \subset \mathbb{R}^{2n}$, $x_1, x_2 \in \mathbb{R}^n$, the state space $X$ is compact, and the controls $u\in U\subset\mathbb{R}^n$. A wide range of robots such as quadrotor and car-like vehicles can be transformed into this form. It is noted that the analysis is restricted to this form, but the results can be extended to the systems of higher relative degree~\cite{khalil2002nonlinear}. In general, $f$ and $G$ may not be fully known for real systems. We make the following assumption:
	
	\noindent\textbf{Assumption 1}\noindent\textbf{:}
	The function $f:X\rightarrow\mathbb{R}^n$ is unknown but has a bounded reproducing kernel Hilbert space (RKHS) norm under a known kernel $k$. The function $G:X\rightarrow\mathbb{R}^{n\times n}$ is known and differentiable, satisfying $Rank(G(x))=n,\ \forall x\in X$. 
	
	The assumption of $f$ limits the irregularity under the induced norm of RKHS, which can be used as a measure for smoothness~\cite{rasmussen2003gaussian}. It is common for practical systems, such as quadrotors and robotic manipulators, that the assumption of $G$ holds such that $G(x)$ is invertible for $x\in X$. 
	
	The goal of this work is to design a novel, learning-based path following control strategy for nonlinear system (\ref{eq:sys_man}), that satisfies the following desired objectives:
	
	\begin{itemize} 
		\item [(R1)] \emph{High-Accuracy Following:} For a given geometric path, the nonlinear system (\ref{eq:sys_man}) moves forward along the path and achieves a high-accuracy path following performance.
		\item [(R2)] \emph{Adaptability:} The proposed strategy can leverage online learning to estimate and adapt to the uncertain environmental disturbances.
		\item [(R3)] \emph{Robustness to Disturbances:} The proposed strategy can optimize the reference target along the path and drive the nonlinear system back to the reference path after irresistible disturbances. \vspace{-1.5mm}
	\end{itemize}

	\subsection{Gaussian Process Regression}
	A Gaussian process (GP)~\cite{rasmussen2003gaussian} is a stochastic process that can be used as a nonparametric regression model to approximate a nonlinear dynamical function, $\delta_i:X\rightarrow\mathbb{R}$, with a fidelity estimation. The GP assumes that function values, associated with different inputs, are random variables and any finite number of them have a joint Gaussian distribution. The approximation of $\delta_i$ can be denoted by 
	\vspace{-1.5mm}
	\begin{equation} 
		\label{eq:one_gp}
		\bar{\delta_i}(x)\sim\mathcal{N}(\mu_i(x), k_i(x, x^\prime)), \vspace{-1.5mm}
	\end{equation}
	which is fully specified by a mean function $\mu_i(x):X\rightarrow\mathbb{R}$ and a kernel $k_i(x, x^\prime):X\times X\rightarrow\mathbb{R}$ estimating the similarity between states $x$ and $x^\prime$. It is common practice to set the prior mean function to zero and use squared-exponential kernel 
	\vspace{-1.5mm}
	\begin{equation}
		k_i(x,\ x^\prime)=\sigma_f^2\exp{(-\frac{1}{2}(x-x^\prime)^\mathrm{T}L^{-2}(x-x^\prime))}, \vspace{-1.5mm}
	\end{equation}
	which is characterized by hyperparameters of the length scale diagonal matrix $L$ and the prior variance $\sigma_f^2$. Since (\ref{eq:one_gp}) represents only functions with a scalar output, $n$ independent GPs can be utilized to model the nonlinear function $\delta:X\rightarrow\mathbb{R}^n$,
	\vspace{-1.5mm}
	\begin{equation}
		\bar{{\delta}}\left(x\right)=\left\{\begin{matrix}{\bar{\delta}}_1\left(x\right)\sim\mathcal{N}(\mu_1\left(x\right),\ k_1\left(x,\ x^\prime\right))\\\ldots\\{\bar{\delta}}_n\left(x\right)\sim\mathcal{N}(\mu_n\left(x\right),\ k_n\left(x,\ x^\prime\right))\\\end{matrix}\right.. \vspace{-1.5mm}
	\end{equation}
	
	We assume that the training set $D$ is available to employ the GPs for regression.

	\noindent\textbf{Assumption 2}\noindent\textbf{:}
	The state $x$ and the function value $\delta(x)$ can be measured with noises over a finite time horizon to make up a training set with $N$ data pairs 
	\vspace{-1.5mm}
	\begin{equation}
		D=\left\{\left(x^{\left(i\right)},\ y^{\left(i\right)}\right)\right\}_{i=1}^N,\ y^{(i)}=\delta\left(x^{(i)}\right)+w_i, \vspace{-1.5mm}
	\end{equation}
	where $w_i$ are i.i.d. noises $w_i\sim N\left(0,\sigma_{noise}^2I_n\right)$, $\sigma_{noise}\in\mathbb{R}$. 
	
	
	Given this training dataset $D$, GPs can be used to predict function value $y^\ast$ at any query input $x^\ast$. The $j-th$ component of the inferred output $y^\ast$ is jointly Gaussian distributed with the training set
	\begin{equation}
		\left[ 
		\begin{array}{c}
			y^\ast_j \\
			{y}_j
		\end{array} 
		\right]\sim \mathcal N(
		\left[ 
		\begin{array}{c}
			0 \\
			{0}
		\end{array} 
		\right], 
		\left[ 
		\begin{array}{cc}
			k^\ast_j & {k}^\mathrm{T}_j\\
			{k}_j & {K}_j + \sigma_{noise}^2 {I}_N
		\end{array} 
		\right]), \vspace{-1.5mm}
	\end{equation}
	where $k^\ast_j=k_j(x^\ast,x^\ast)\in\mathbb{R}$, ${y}_j=\left[y^{(1)}_j,\ldots,y^{(N)}_j\right]^\mathrm{T}\in\mathbb{R}^N$, ${k}_j=\left[k_j\left(x^{\left(1\right)},x^\ast\right),\ \ldots,k_j\left(x^{\left(N\right)},x^\ast\right)\right]^\mathrm{T}\in\mathbb{R}^N$, and ${K}_j\in\mathbb{R}^{N\times N}$ is covariance matrix with the entry $\left[{K}_j\right]_{(l,m)}=k_j\left(x^{\left(l\right)},x^{\left(m\right)}\right),\ l,\ m=1,\ldots,N$. 
	
	Conditioning on the test input $x^\ast$ and the training data $D$, the distribution on $\bar{\delta}\left(x^\ast\right)$ is $\bar{\delta}\left(x^\ast\right)\sim\mathcal{N}(\mu(x^\ast), \sigma^2(x^\ast))$ with the $j-th$ component of the mean and variance:
	\begin{equation}
		\begin{split}
			\mu_j\left(x^\ast\ \right)&={k}^\mathrm{T}_j\left({K}_j+\sigma_{noise}^2{I}_N\right)^{-1}{y}_j,\\ 
			\sigma^2_j\left(x^\ast\ \right)&=k^\ast_j-{k}^\mathrm{T}_j\left({K}_j+\sigma_{noise}^2{I}_N\right)^{-1}{k}_j.
		\end{split}
	\end{equation}

	With the function $\bar{\delta}$ approximated by GPs, the following result allows us to quantify the upper bound of the difference between the true function $\delta(x)$ and the inferred mean $\mu(x)$ with a reliable confidence interval. 
	
	\noindent\textbf{Lemma 1} (\hspace{-0.05mm}\cite{umlauft2018uncertainty})\noindent\textbf{:}
	For any compact set $X\subset\mathbb{R}^{2n}$ and a probability $\varsigma\in(0,\ 1)$ holds
	\begin{equation}
		\label{eq:prob}
		Pr\{\| \mu(x)-\delta(x) \| \leq \|\beta\|\|\sigma(x)\|, \forall x \in X \} \geq (1-\varsigma)^{2n}, 
	\end{equation}
	where $Pr$ denotes probability, $\beta = [\beta_1, ..., \beta_n]^\mathrm{T}$, $\beta_j=(2\|{\delta_j\|^2}_{k_j}+300\gamma_j ln^3(\frac{N+1}{\delta}))^{\frac{1}{2}}$, $j=1,...,n$, $\gamma_j$ is the maximum information gain under the kernel $k_j$:
	$\gamma_j = \max\limits_{\{x^{\left(1\right)},\ \ldots,\ x^{(N)}\}\in X}\frac{1}{2}\log(\det(I_N - \sigma^{-2}_{noise}{K}_j(x, x{\prime})))$, 
	$x, x^{\prime} \in \{x^{\left(1\right)},\ \ldots,\ x^{(N)}\}$, .
  
	\textbf{Lemma 1} allows us to make high probability statements on the maximum modeling error between the true function $\delta$ and the inferred mean $\mu$, and it will be utilized in the analysis and synthesis of the proposed control scheme. 

	\section{Methodology}
	\label{sec:alg}
	To achieve the three goals in \ref{sec:problem statement}, we propose a hierarchical control scheme composed of a high-level MPFC coupled with an underlying LB-FBLC, as shown in Fig. \ref{fig:structure}. We first introduce the LB-FBLC in Section~\ref{sec:alg}-A and the high-level MPFC in Section~\ref{sec:alg}-B. 	
	\vspace{-1mm}
	\subsection{Learning-based Feedback Linearization Controller (LB-FBLC)}
	Suppose we are given a bounded reference state $x_d(t)=\left[x_{1d}(t),x_{2d}(t)\right]^\mathrm{T}\in X$ and a control $a_d(t)\in \mathcal{A} \subset \mathbb{R}^n$ from the high-level MPFC. Since the MPFC uses an integrator as the predictive model, these references satisfy 	
	\vspace{-1mm}
	\begin{equation}
		{\dot{x}}_{1d}(t)=x_{2d}(t),\ \ {\dot{x}}_{2d}(t)=a_d(t). 
		\label{eq:desired_integrator}	\vspace{-1mm}
	\end{equation}
	For simplicity, we omit the time index in the following.

	A common method to track the reference state for nonlinear systems (\ref{eq:sys_man}) is feedback linearization control. Since $f\left(x\right)$ is not known exactly, let $\hat{f}(x)$ be a nominal model of $f(x)$. We formulate the feedback linearization control law $u$, with pseudo-control component $a$, 	
	\vspace{-1mm}
	\begin{equation}
		u=G\left(x\right)^{-1}(a-\hat{f}(x))
		\label{eq:inverse_control}	\vspace{-1mm}
	\end{equation}
	to convert the nonlinear system (\ref{eq:sys_man}) into an approximately linear integrator model	
	\vspace{-1mm}
	\begin{equation}
		\dot{x}_1=x_2,\ \dot{x}_2=a+\delta\left(x\right), 
		\label{eq:integrator} \vspace{-1mm}
	\end{equation}
	where $\delta\left(x\right)=f\left(x\right)-\hat{f}(x)\in\mathbb{R}^n$ is the modeling error resulting from environmental disturbances. If the nominal model matches the actual model, then $\delta\left(x\right)=0$ and (\ref{eq:integrator}) becomes a double integrator. However, it is difficult to get an accurate model in advance for practical robotic systems under uncertain environmental disturbances.
	
	To achieve precise tracking control given reference state $x_d$ and control $a_d$, we design the pseudo-control $a$ as
	\begin{equation}
		a=a_d+K_P\left(x_{1d}-x_1\right)+K_D\left(x_{2d}-x_2\right)+r,
		\label{eq:control_all}
	\end{equation}
	where $K_P\in\mathbb{R}^{n\times n}$, $K_D\in\mathbb{R}^{n\times n}$ are the proportional and derivative matrices of PD control law~\cite{khalil2002nonlinear}, respectively, and $r\in\mathbb{R}^n$ is an added vector to be designed to compensate the disturbances and achieve tracking stability.

	Define the tracking error $e=x-x_d$. Then, it can be shown from (\ref{eq:integrator}) and (\ref{eq:control_all}) that the tracking error dynamics can be written as	
	\vspace{-1mm}
	\begin{equation}
		\dot{e}=Ae+B\left(r+\delta\left(x\right)\right),
		\label{eq:error_dyn}\vspace{-1mm}
	\end{equation}
	where $A=\left[\begin{matrix}0&I_n\\-K_P&-K_D\\\end{matrix}\right]\in\mathbb{R}^{2n\times2n}$, and $B=\left[\begin{matrix}0\\I_n\\\end{matrix}\right]\in\mathbb{R}^{2n\times n}$. The control gain matrix $K_P$, $K_D$ should be chosen to make $A$ a Hurwitz matrix~\cite{khalil2002nonlinear}. Let $P\in\mathbb{R}^{2n\times2n}$ be the unique positive definite matrix, satisfying $A^\mathrm{T}P+PA=-Q$, where $Q\in\mathbb{R}^{2n\times2n}$ is a positive definite matrix.
	
	We now discuss how to design the adaptive control vector $r$. Intuitively, the environmental disturbances can be totally compensated when they are known exactly, i.e. $r=-\delta(x)$. However, only an approximation $\bar{\delta}$ of $\delta(x)$ can be obtained from limited data. Thus, the approximation methods should provide the estimation of $\delta(x)$ and capture the uncertainty of the estimation. To this end, GPs are utilized to predict the environmental disturbances and quantify the uncertainty based on its predictions, where the output of the GPs is $\bar{\delta}\left(x\right) \sim\mathcal{N}(\mu\left(x\right), \sigma(x))$ as illustrated in Section~\ref{sec:problem}. To train the GPs, $N$ data points can be collected online to makeup the training dataset $D=\{(x^{(i)},\ {(\dot{x}_2-a)\ }^{(i)})\}_{i=1}^N$.

	With the estimated disturbance $\bar{\delta}\left(x\right)$, the adaptive control vector $r$ can be designed as 
	\vspace{-1mm}
	\begin{equation}
		r=-\mu\left(x\right)-k_cB^\mathrm{T}Pe,
		\label{eq:control_r}\vspace{-1mm}
	\end{equation} 
	where $k_c\in\mathbb{R}$ is an adjustable control parameter.
	
	\noindent\textbf{Lemma 2}\noindent\textbf{:}
	Consider the system (\ref{eq:sys_man}) with a bounded desired state $x_d$. Suppose that \textbf{Assumptions 1} and \textbf{Assumptions 2} hold. Then, the proposed learning-based control strategy in (\ref{eq:inverse_control}), (\ref{eq:control_all}) and (\ref{eq:control_r}) with the condition
	\vspace{-1mm}
	\begin{equation}
		k_c\|B^\mathrm{T}Pe(x)\| - \|\beta\|\|\sigma(x)\|\geq 0, \forall x \in X,
		\label{eq:condition}	\vspace{-1mm}
	\end{equation}
	ensures that the tracking error $e$ semi-globally asymptotically converges to zero with probability at least $(1-\varsigma)^n$ for $e\in \mathcal{E}$, where $\mathcal{E}$ is a compact set $\mathcal{E}=\{e\in\mathbb{R}^{2n}|e^\mathrm{T}Pe\le e\left(0\right)^\mathrm{T}Pe(0)\}$.
	
	\begin{IEEEproof}
		Consider a candidate Lyapunov function $V(e)=e^\mathrm{T}Pe$. Denote $w=w^\mathrm{T}=B^\mathrm{T}Pe$. It can be shown from (\ref{eq:error_dyn}) that $\dot{V}(e)=-e^\mathrm{T}Qe+2w(r+\delta)$. With control law (\ref{eq:control_r}), we have 
		\vspace{-1mm}
		\begin{equation}
			\begin{aligned}
				\dot{V}(e)&= -e^\mathrm{T}Qe+2w(\delta-\mu-k_cw) \\
				&\leq -e^\mathrm{T}Qe+2w(\delta-\mu)-2k_c\|w\|^2 \\
				&\leq -e^\mathrm{T}Qe+2\|w\|\|\delta- \mu\|-2k_c\|w\|^2,
			\end{aligned} \vspace{-1mm}
		\end{equation}
		where the inequality comes from the Cauchy-Schwarz inequality. Employing \textbf{Lemma 1}, we have $Pr\{\dot{V}(e)\leq -e^\mathrm{T}Qe + 2 \|w\|(\|\beta\|\|\sigma(x)\| - k_c\|w\|), \forall e \in \mathcal{E}\} \geq (1-\varsigma)^n$. It yields $Pr\{\dot{V}(e)<0, \forall e \in \mathcal{E}\verb|\|\{0\} \} \geq (1-\varsigma)^n$ under the condition (\ref{eq:condition}). This strict inequality holds because $-e^\mathrm{T}Qe<0$ with $Q$ a positive definite matrix. In addition, $\dot{V}(0)=0$ holds.
	\end{IEEEproof}
	
	Nonlinear system (\ref{eq:sys_man}) is forced to behave like an integrator (\ref{eq:desired_integrator}) leveraging the control law (\ref{eq:control_r}). It is noted that if the variance $\sigma\left(x\right)=0$, i.e. the prediction of disturbances $\bar{\delta}(x)$ via GPs matches the true value $\delta(x)$ ideally, the tracking error $e$ will asymptotically converge to zero and the system (\ref{eq:sys_man}) can be transformed exactly into an integrator.
	
	Additionally, considering the control constraints, we can leverage the control Lyapunov function to construct a quadratic programming (QP) to obtain the parameter $k_c$: 
	\begin{alignat}{2}
		\label{opt_1}
		k_c^{*}=\arg\min_{k_c} \quad & \|k_cw\|_2^2 + k_{\epsilon}\epsilon^2,\\
		\mbox{s.t.} \quad
		&H_{clf}k_c + b_{clf} \leq\epsilon,\tag{Stability Constraints} \\
		&H_{u}k_c + b_{u} \leq 0,\tag{Control Constraints}
	\end{alignat}
	where $H_{clf}=-2\|w\|^2$, $b_{clf}=-e^\mathrm{T}Qe+2\|w\|\|\beta\|\|\sigma(x)\|$, $H_u=[G(x)^{-1}w,\ -G(x)^{-1}w]^\mathrm{T}$, $b_u=[G(x)^{-1}(\mu-a_d-a_{pd}+\hat{f}(x))+u_{min}, G(x)^{-1}(-\mu+a_d+a_{pd}-\hat{f}(x))-u_{max}]^\mathrm{T}$, $a_{pd}=K_P\left(x_{1d}-x_1\right)+K_D\left(x_{2d}-x_2\right)$, $\epsilon \in \mathcal{R}$ is a slack variable to ensure the QP is feasible.	
	
	\noindent\textbf{Remark}\noindent\textbf{.} Note that the optimization~(\ref{opt_1}) is not sensitive to the $k_{\epsilon}$ parameter as long as it is large enough (e.g. $10^{20}$), such that stability constraints violation is heavily penalized. 
	\subsection{Model Predictive Path Following Control (MPFC)}	
	With the integrator (\ref{eq:desired_integrator}) as the predictive model, an MPFC is designed to optimize the reference target along the path, and provide reference state $x_d$ and reference control $a_d$ for the LB-FBLC. As a high-level controller, it accounts for disturbances by optimizing the speed of reference target evolution along the path in a framework of nonlinear model predictive control (NMPC). 
	 
	Given a geometric path $\mathcal{P}$:
	\vspace{-1mm}	
	\begin{equation}
		\label{eq:path}
		\mathcal{P}=\{x_{ref}\in{X\subset\mathbb{R}}^{2n} \lvert x_{ref}=P(\theta),\ \theta\in\Theta\},	\vspace{-1mm}
	\end{equation}
	which is described by a projection $P:\Theta\rightarrow X$, from a parameter interval $\Theta=[\theta_0,\theta_{end}]\subset \mathbb{R}$ to the state space $X$.
	This parameter can be regarded as an additional degree of freedom to be optimized in the NMPC. However, changing $\theta(t)$ directly can result in undesired jumps along the path. To avoid this effect, a virtual dynamics model of reference target is designed as another integrator:
	\vspace{-1mm}
	\begin{equation}	
		\dot{\theta}=\theta_{vel}, \ 
		{\dot{\theta}}_{vel}=\theta_{acc},
		\label{eq:theta_dyn}	\vspace{-1mm}
	\end{equation}
	in which $\theta_{vel}>0$, $\theta_{acc}\in\mathcal{A}_{\Theta}\subset\mathbb{R}$ is the control input to the virtual system (\ref{eq:theta_dyn}) and $\mathcal{A}_{\Theta}$ is a compact set containing the origin in its interior. This control input $\theta_{acc}$ can be regarded as the acceleration of the reference target evolution. The relative degree of virtual system is designed corresponding to the linearized system (\ref{eq:integrator}) and can provide smooth evolution of $\theta$ by optimizing its acceleration.~\cite{faulwasser2015nonlinear} gives a differential algebraic and geometric explanation of this kind of virtual reference target dynamics. 
	
	Based on the linearized system dynamics (\ref{eq:desired_integrator}) and the linear virtual target dynamics (\ref{eq:theta_dyn}), a continuous time sampled-data NMPC scheme is constructed. As commonly in the MPC, the system input is obtained by repetitively solving a finite-time optimal control problem (OCP). Denote the sampling instance $t_k=t_0+k\cdot dt$, where $k\in\mathbb{N}$, $t_0$ is the initial time instant, and $dt$ is the control period. Specifically, at each sampling instance $t_k$, the following OCP is solved: 
	\vspace{-1mm}
	\begin{alignat}{2}
		\label{opt_2}
		\min_{\bar{a}(t),\ \bar{\theta}_{acc}(t)} \quad & \ J\left(x(t_k),\ \theta(t_k),\bar{a}(t),\ \bar{\theta}_{acc}(t)\right),\\
		\mbox{s.t.} \quad
		&\dot{\bar{x}}_1(t)={\bar{x}}_{2}(t),\ \ \ \ \dot{\bar{x}}_2(t)=\bar{a}(t), \\
		&\dot{\bar{\theta}}(t)={\bar{\theta}}_{vel}(t),\ \ \ \ {\dot{\bar{\theta}}}_{vel}(t)=\bar{\theta}_{acc}(t),\\
		&{\bar{\theta}}_{vel}(t)>0,\ \ \ \label{eq:theta_dot} \\
		&\bar{x}(t_k)=x(t_k),\ \ \ \ \bar{\theta}(t_k)=\theta(t_k),\\
		&\bar{x}\left(t\right)\in X,\ \ \ \ \ \ \ \ \ \bar{a}(t)\in \mathcal{A},\\
		&\bar{\theta}\left(t\right)\in\Theta,\ \ \ \ \ \ \ \ \ \ \bar{\theta}_{acc}(t)\in\mathcal{A}_{\Theta}, \label{opt_2_end} \vspace{-1mm}
	\end{alignat}
	where the superscript $\bar{.}$ denotes the predicted state or control variables. 
	 
	The objective function $J$ at $t_k$ can be designed as follows.
	\vspace{-1mm}
	\begin{equation}
		\small
		\begin{aligned}
			\label{eq:object}
			J(x(t_k),\theta(t_k),\bar{a}(t),\bar{\theta}_{acc}(t))
			= &\int_{t_k}^{t_k + H\cdot dt} \|\bar{x}(t)-P(\bar{\theta}(t))\|^2_{Q_{mpc}} \\& +\|\bar{a}(t)\|^2_{R_{a}} + R_{\theta}\bar{\theta}_{acc}(t)^2 dt,
		\end{aligned} \vspace{-1mm}
	\end{equation}
	where $H$ is the predictive steps, positive semi-definite matrix $Q_{mpc}\in\mathbb{R}^{2n\times2n}$ weights the tracking error between the system states and reference targets, and positive definite matrix $R_{a\ }\in\mathbb{R}^{n\times n}$ and $R_\theta>0$ ensure regularization of the inputs. Notice that constraint (\ref{eq:theta_dot}) guarantees forward moving of reference target along the path.
	
	The solution to the OCP (\ref{opt_2})-(\ref{opt_2_end}) contains reference target $P({\bar{\theta}}^\ast(t))$, state ${\bar{x}}^\ast(t)$ and control input ${\bar{a}}^\ast(t)$, $\forall t\in [t_k,t_k+dt]$. The reference and control are applied to the LB-FBLC as $x_d(t)$ and $a_d(t)$. At the next sampling instant $t_{k+1}=t_k+dt_{mpc}$, the OCP (\ref{opt_2})-(\ref{opt_2_end}) will be solved again with new measured states served as an initial condition. As discussed in~\cite{faulwasser2015nonlinear}, the path convergence and recursive feasibility in the presence of system constraints can be ensured by carefully adding an end penalty and a terminal constraint to the OCP (\ref{opt_2})-(\ref{opt_2_end}). 
	\vspace{-1mm}
	\section{APPLICATION TO QUADROTOR AND SIMULATION RESULTS}
	\vspace{-1mm}
	\label{sec:sim}
	In this section, the proposed approach is applied to a quadrotor to follow different predefined paths in the presence of unknown wind disturbances. Predictive following performance and adaptability to the uncertain disturbances of the proposed framework are verified via simulation.
	\vspace{-1mm}
	\subsection{Quadrotor Dynamics and Control}
	\vspace{-1mm}
	The quadrotor is a well-modeled dynamic system with torques and forces generated by four rotors and gravity. The Euler angles (roll $\phi$, pitch $\theta$ and yaw $\psi$) are defined with the ZYX convention. Thus, the attitude rotation matrix $R\in SO(3)$ from the body frame $B$ to the world frame $W$ can be written as
	\begin{equation}
		R=\left[\begin{matrix}c\theta c\psi&s\phi s\theta c\psi-c\phi s\psi&c\phi s\theta c\psi+s\phi s\psi\\c\theta s\psi&s\phi s\theta s\psi+c\phi c\psi&c\phi s\theta s\psi-s\phi c\psi\\-s\theta&s\phi c\theta&c\phi c\theta\\\end{matrix}\right],
	\end{equation}
	where $s$ and $c$ represents $sin$ and $cos$, respectively~\cite{hehn2015real}.
	
	Given quadrotor states as position $p\in\mathbb{R}^3$ and velocity $v\in\mathbb{R}^3$ in the world frame $W$, we consider the following translational dynamics:
	\vspace{-1mm}
		\begin{equation}
		\begin{aligned}
		\label{eq:quad_dyn}
		\dot{p}&=v, \\	
		\dot{v}&=-ge_3+\frac{1}{m}Rf_u+\frac{1}{m}f_a,
		\end{aligned}	
	\vspace{-1.5mm}
		\end{equation}
	where $m$ is the mass of the quadrotor, $e_3=[0,0,1]^T$ is the unit vector, $g$ is the gravitational acceleration, and $f_u=[0,0,f_T]^\mathrm{T}$ with $f_T$ the total thrust generated from four rotors. The wind disturbance is $f_a=K_{drag}(v_w-v)$, where $v_w\in\mathbb{R}^{3}$ is the velocity of wind disturbances in $W$ and $K_{drag}\in\mathbb{R}^{3\times3}$ is a drag coefficient diagonal matrix.
	We define in the dynamics equation (\ref{eq:sys_man}) the state $x=[x_1, x_2]^\mathrm{T}=[p,v]^\mathrm{T}$, with $x_1=p=[p_x,p_y,p_z]^\mathrm{T}\in\mathbb{R}^3$ and $x_2=v=[v_x,v_y,v_z]^\mathrm{T}\in\mathbb{R}^3$, and define the total desired rotor force as control $u=(Rf_u)_d\in\mathbb{R}^3$ following~\cite{shi2019neural}. 

		\begin{figure}[t]
			\centering
			\includegraphics[scale=0.12]{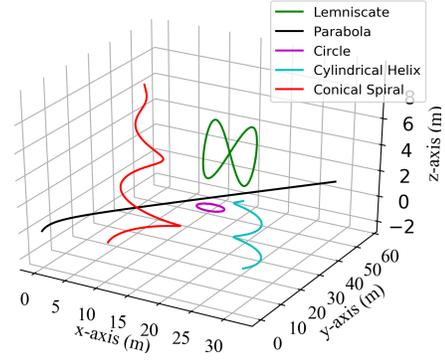}
			\caption{The five different paths used to test the proposed path following control scheme.}
			\vspace{-7mm}
			\label{fig:all_paths}
		\end{figure}
	
	We assume that the attitude controller is provided by a commercial quadrotor with the control interface ($\phi_{cmd}$, $\theta_{cmd}$, $\psi_{cmd}$, $f_{Tcmd}$). Given geometric paths as described in \ref{sec:setup}, reference states $x_d$ and controls $a_d$ are provided by the MPFC. With these references, the control $u=\left[u_x,u_y,u_z\right]^\mathrm{T}$ as well as $a=\left[a_x,a_y,a_z\right]^\mathrm{T}$ are computed using $(\ref{eq:inverse_control}), (\ref{eq:control_all})$ and $(\ref{eq:control_r})$ with $k_c$ computed by solving ($\ref{opt_1}$). As shown in Fig. \ref{fig:structure}, an inner-loop controller leveraging the differential flatness property of the quadrotor~\cite{mellinger2011minimum} converts the computed controls to the attitude and thrust commands. Specifically, these commands can be computed as follows~\cite{zhou2014vector}.
	\vspace{-0.5mm}
	\begin{equation}
		\begin{aligned}
			f_{Tcmd}=\|u\|,\  \theta_{cmd}=atan2\left(\beta_a,\ \beta_b\right),\\
			\phi_{cmd}=atan2\left(\beta_c,\ \sqrt{\beta_a^2+\beta_b^2}\right), \vspace{-2.5mm}
		\end{aligned}
	\end{equation}
	where $\beta_a=-a_x\cos{{\psi}_{cmd}}-a_y\sin{{\psi}_{cmd}}$, $\beta_b=-a_z+g$, and $\beta_c=-a_x\sin{{\psi}_{cmd}}+a_y\cos{{\psi}_{cmd}}$. The command of yaw is set as $\psi_{cmd}=0$.
	\vspace{-2mm}
	\subsection{Simulation Setup}
	\label{sec:setup}
	\vspace{-1mm}
	We create a simulation platform using Python 3.7 to numerically validate the performance of the proposed control methodology for quadrotor following paths under unknown wind disturbances. The quadrotor model corresponds to the commercial quadrotor Crazyflie 2.1 with mass $m=0.036kg$. The control constraints of $u$ are set $u_{min}=m\cdot[-2, -2, -2+g]^T (N)$ and $u_{max}=m\cdot[2, 2, 2+g]^T (N)$ \cite{mueller2013model}. The drag coefficient is set $K_{drag}=diag[0.02, 0.02, 0.02]$.
	The proposed path following controller runs at 100 $Hz$ with a control period $dt=0.01s$. The internal prediction steps of the MPFC is $H=20$. The weights in the objective function (\ref{eq:object}) of NMPC are set as $Q_{mpc}=diag[10,10,10,1,1,1]$, $R_{a}=diag[0.1,0.1,0.1]$ and $R_{\theta}=0.1$.  
		
 \begin{table}[tp]
	\scriptsize
	\centering
	\renewcommand{\arraystretch}{1.1}
	\caption{Averaged Path Following RMSEs (in meter) over the Five Paths for Different Control Schemes} 
	\vspace{-2mm}
	\label{table:adaptive_table}
	\begin{tabular}{cc|cc}
		\hline
		High-level & Low-level    & 
		\begin{tabular}{@{}c@{}} Without\\Disturbances \end{tabular}  & 
		\begin{tabular}{@{}c@{}} \textbf{Uncertain} \\ \textbf{Disturbances} \end{tabular} \\
		\hline
		MPFC            & FBLC          & 0.0190 & 0.0346\\
		\textbf{MPFC}   & \textbf{LB-FBLC} & \textbf{0.0144 } & \textbf{0.0162 } \\
		MPFC            & FFLC            & 0.0353 & 0.0376 \\
		MPFC            & LB-FFLC         & 0.0380  & 0.1102\\
		MPFC            & ROBUST~\cite{helwa2019provably}         & 0.0168 & 0.0368 \\
		\hline
		Carrot-chasing         & LB-FBLC          & 0.1580  & 0.1653 \\
		NLGL            & LB-FBLC         & 0.1692  & 0.1785\\
		\hline
	\end{tabular} 
\vspace{-2mm}
\end{table}

	\begin{table}[tp]
		\renewcommand{\arraystretch}{1.1}
		\scriptsize
		\caption{Maximum Following Errors (in meter) of a quadrotor following 5 Paths with Different high-level Controllers}
		\label{table:table_predictivity}
		\centering
		\vspace{-2mm}
		\begin{tabular}{ c | c c c c c }
			\hline
			High-level & \thead{Lemniscate}& \thead{
				Parabola} & \thead{Circle} & \thead{CH} & \thead{CS}\\
			\hline
			MPFC & \textbf{0.5667} & \textbf{0.2221}  & \textbf{0.1384}  & \textbf{0.8190} & \textbf{0.4397}\\
			Carrot-chasing & 0.6050 & 0.3299 & 0.3637  & 1.4188 &0.5945 \\
			NLGL & 0.6020 & 0.3290  & 0.3868  & 1.3200 & 0.6638\\
			\hline
		\end{tabular}	
	\vspace{-4mm}
	\end{table}
	\begin{figure}[tp]  
		\centering 
		\subfigure[]{
			\label{fig:GP}
			\includegraphics[scale=0.065]{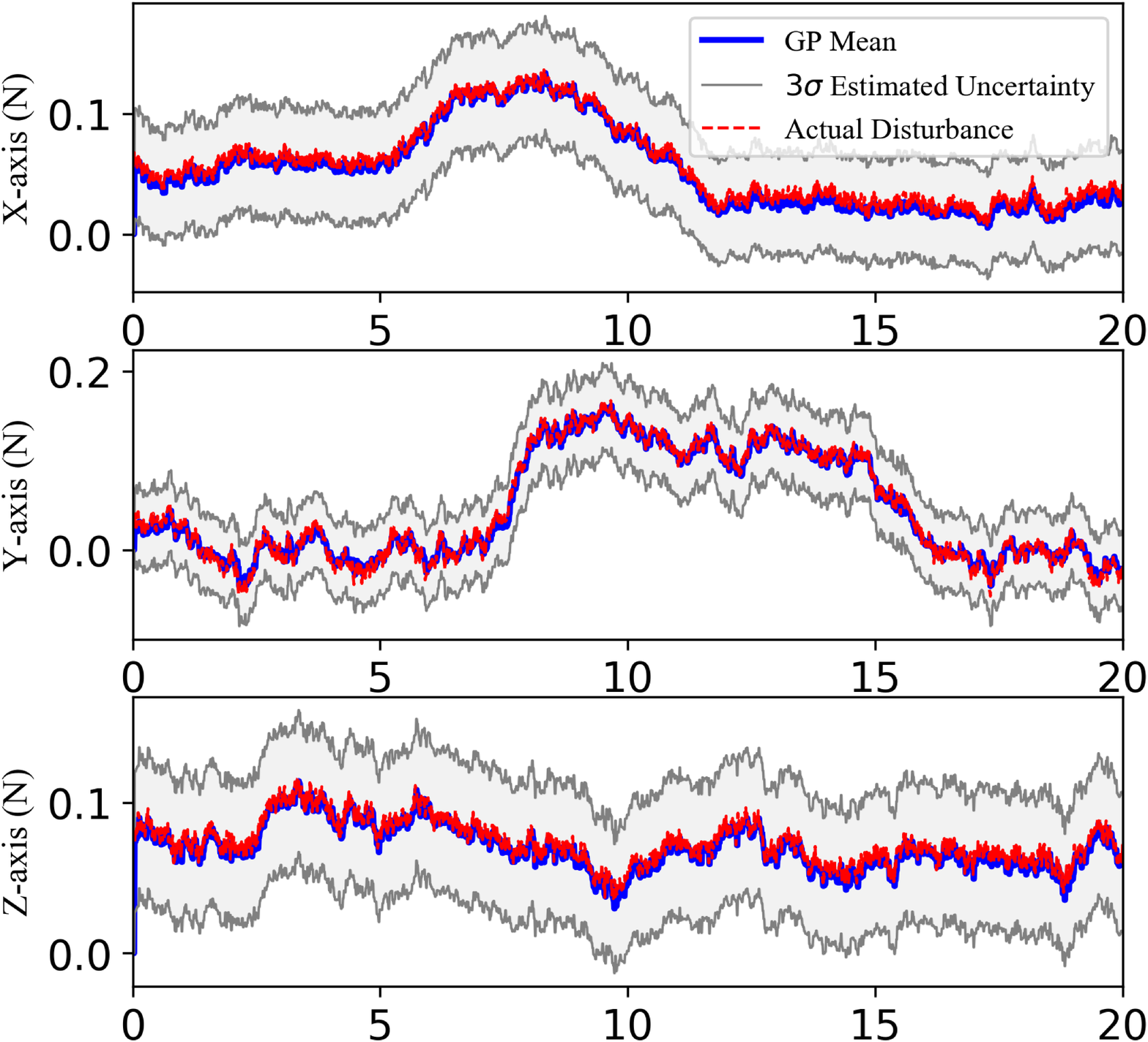}}
		\subfigure[]{
			\label{fig:linear_integrator}
			\includegraphics[scale=0.065]{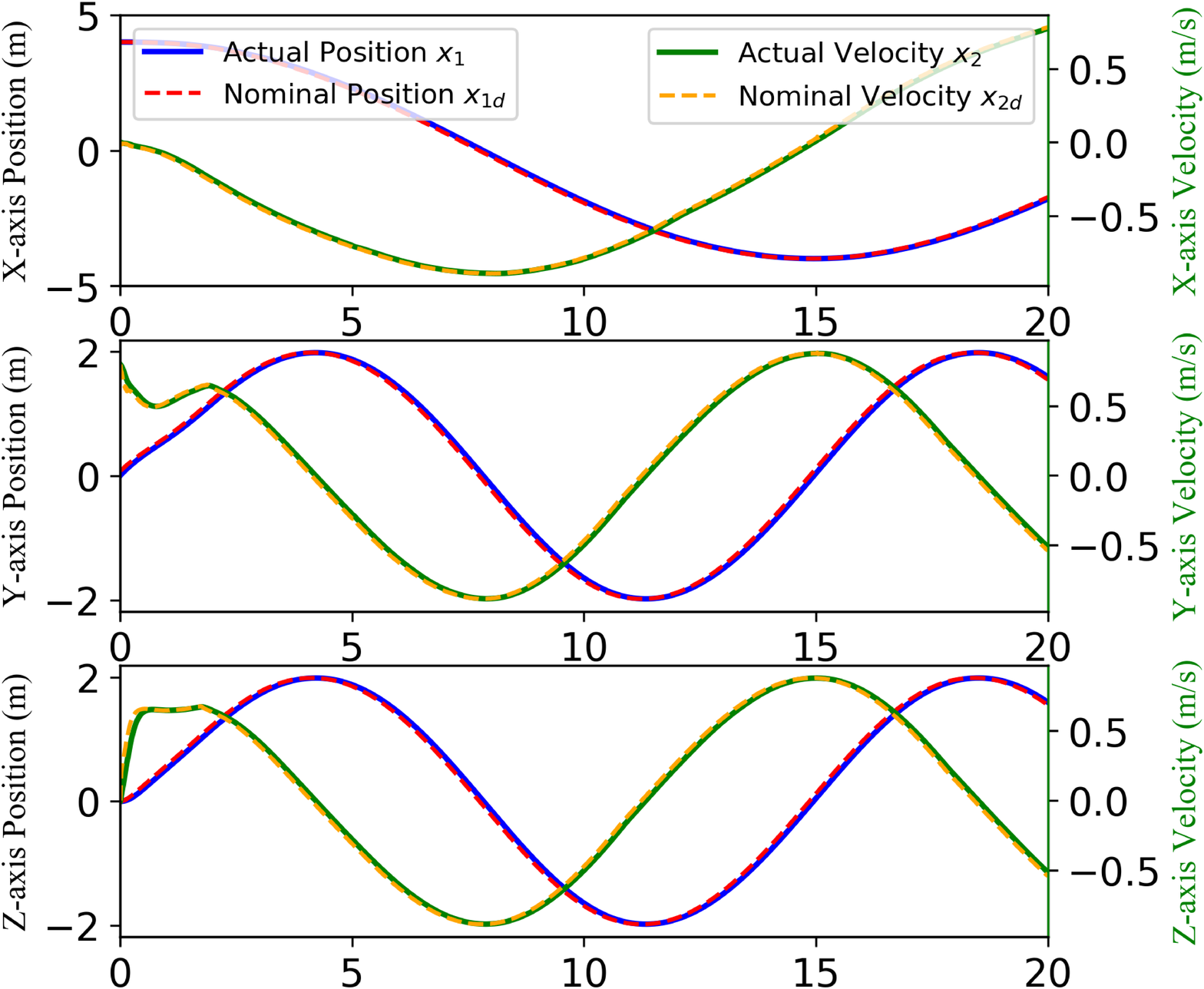}}
		\vspace{-2mm}
		\caption{(a) The estimated disturbances in three axes via GPs when following the lemniscate path. (b) The nominal trajectory followed
			by the ideal system (9) is compared to the actual trajectory followed by the quadrotor.}
		\label{fig:integrator}  
		\vspace{-2mm}
	\end{figure}

	\begin{figure}[tp]
		\centering
		\includegraphics[scale=0.12]{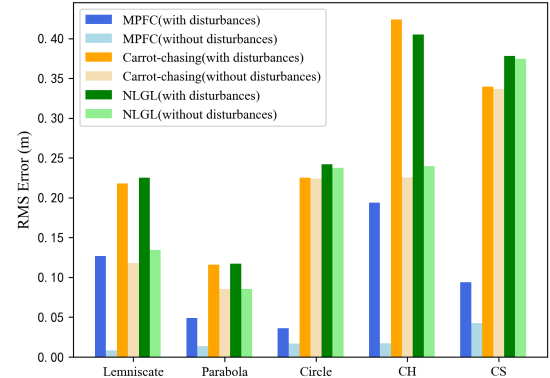}	
		\vspace{-2.5mm}
		\caption{The RMSEs of the quadrotor following five different paths using different high-level controllers without disturbances and under the same wind disturbances.}	
		\vspace{-6mm}
		\label{fig:trajs}
	\end{figure}
	The control gain matrix in PD control is $K_p=diag[2, 2, 2]$ and $K_d=diag[1, 1, 1]$. The Q matrix in the Lyapunov function is set as $Q=diag[1, 1, 1, 1, 1, 1]$. The penalty coefficients of the slack variable in the QP are set $k_{\epsilon}=1e20$. The QP is solved with the OSQP solver~\cite{osqp}. We use the scikit-learn Python package~\cite{pedregosa2011scikit} to build 3 GPs to estimate unknown wind disturbances $f_a$. Each GP uses the same squared-exponential kernel with parameters $L=10$ and $\sigma_f=1$. The GPs update with the past $N=5$ data pairs as the training set for prediction. We set $\beta=3$ in (\ref{eq:prob}) for each GP. The computing time for GP learning is $0.1s$ and for inference is below $0.003s$ on average on a 2.10GHz Intel Xeon CPU. Solving the NMPC in the simulation takes $0.07s$ on average using the CasADi ~\cite{andersson2019casadi} with the IPOPT solver \cite{biegler2009large} in Python. These codes have not been optimized for speed and can be much accelerated in C++.

	\begin{figure*}[tp]  
		\centering  
		\hspace{-1cm}
		\subfigure[MPFC]{
			\label{fig:position_error1}
			\includegraphics[scale=0.024]{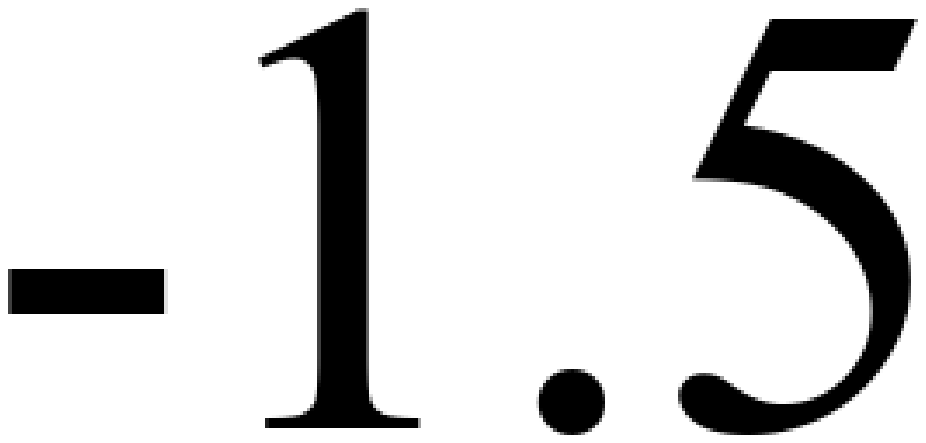}}\hspace{2mm}
		\subfigure[Carrot-chasing]{
			\label{fig:position_error2}
			\includegraphics[scale=0.024]{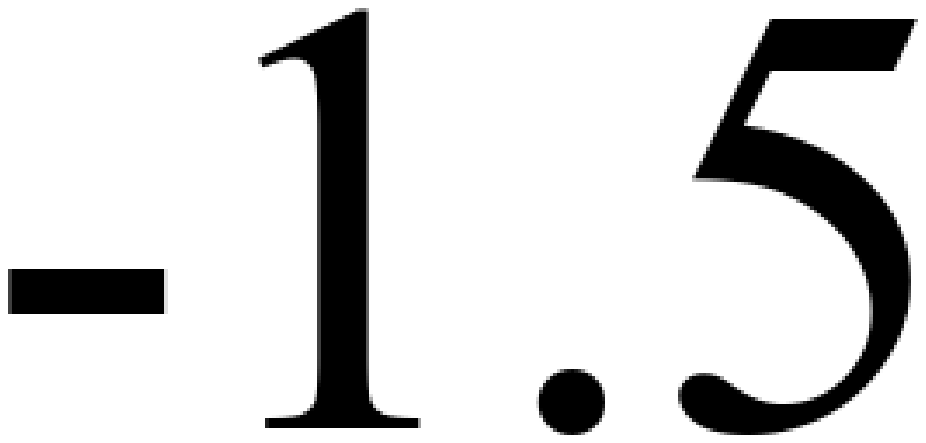}}\hspace{2mm}
		\subfigure[NLGL]{
			\label{fig:position_error3}
			\includegraphics[scale=0.024]{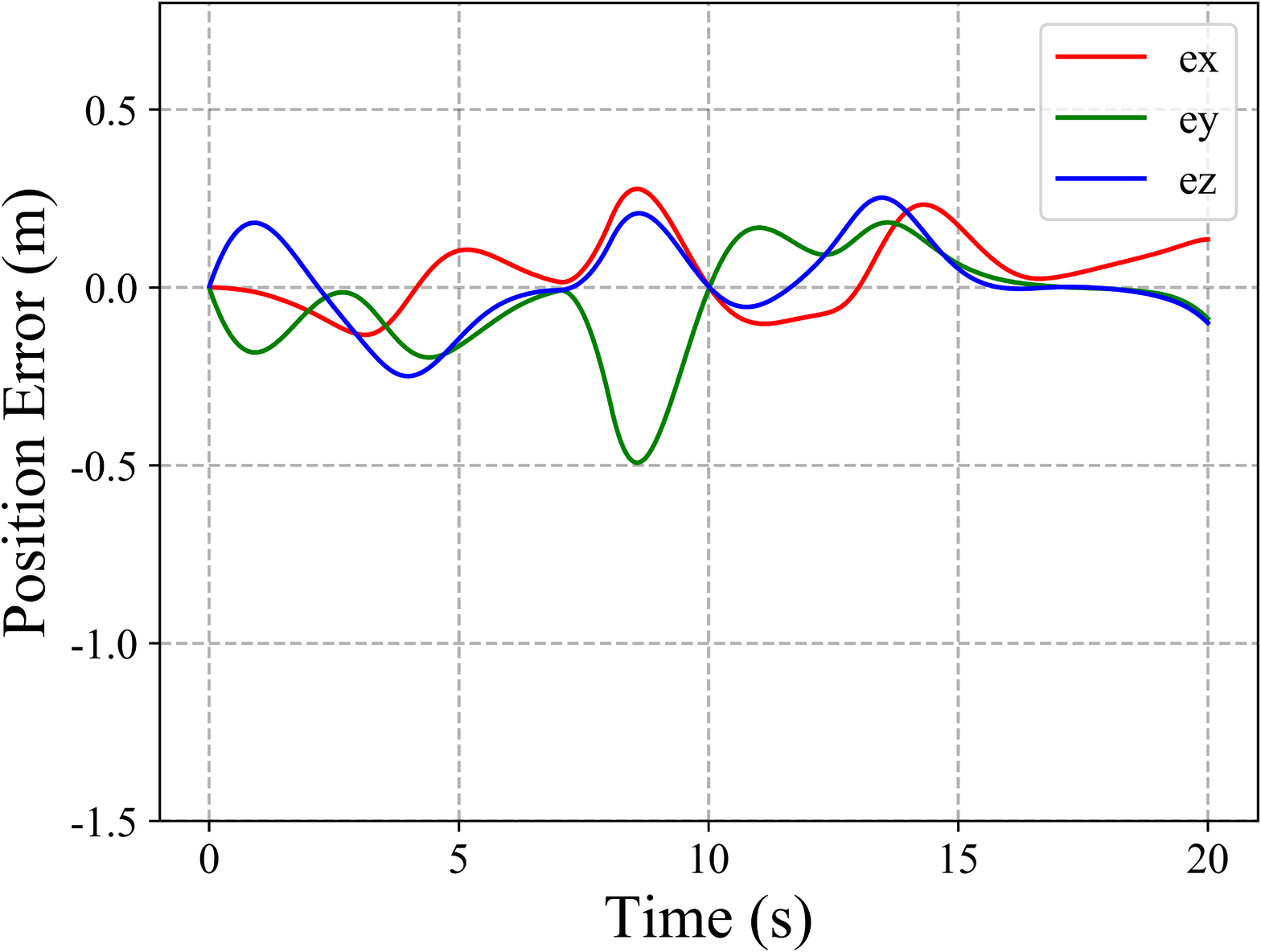}}\hspace{2mm}
		\subfigure[MPC]{
			\label{fig:mpc}
			\includegraphics[scale=0.024]{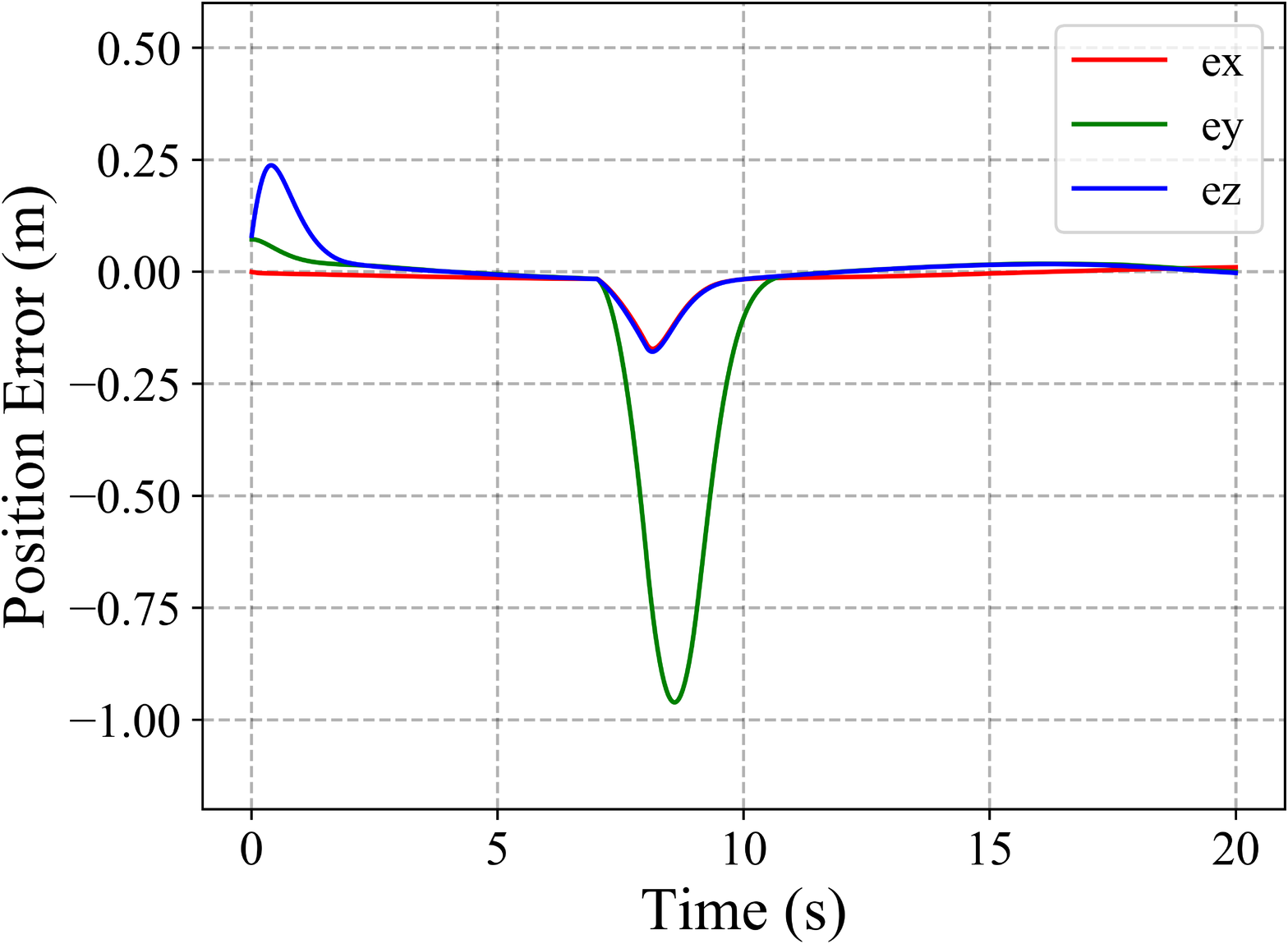}}
		\vspace{-3mm}
		\caption{Position error of quadrotor following the aggressive lemniscate path under wind disturbances. Note that the deviation at the beginning is caused by wind disturbances, which has not been estimated and compensated by GPs with limited data. The jumps at around 7$s$ are caused by an irresistible wind gust. The proposed control scheme with MPFC as the high-level controller outperforms the other three baseline controllers in terms of position error and overshooting.} 
		\label{fig:disturbance_error} 
		\vspace{-5mm}
	\end{figure*}
 
	\begin{figure}[tp]
		\centering  
		\subfigure[Path Parameter $\theta$]{
			\label{fig:theta_v}
			\includegraphics[scale=0.025]{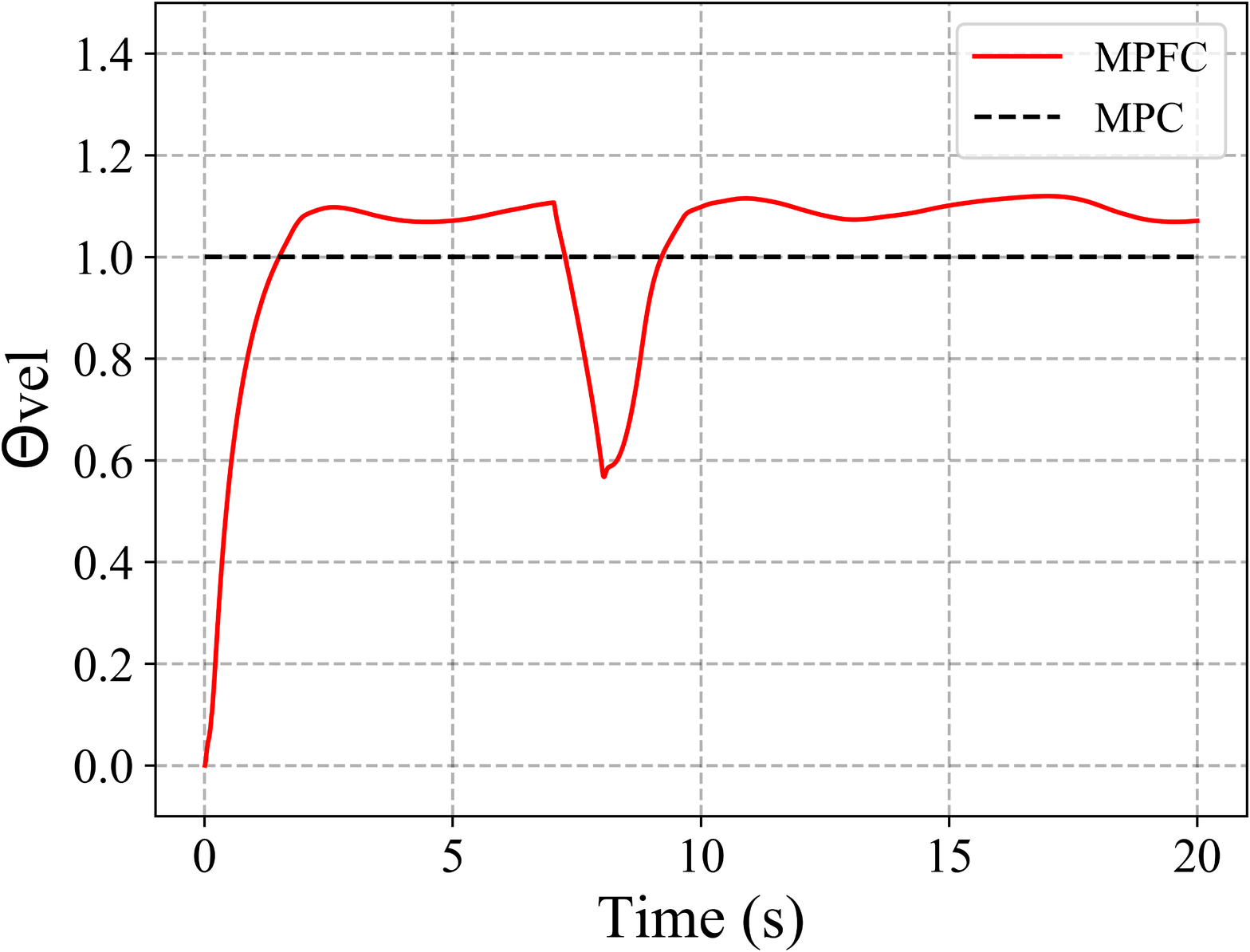}}
		\subfigure[Control Inputs]{
			\label{fig:control_Inputs}
			\includegraphics[scale=0.025]{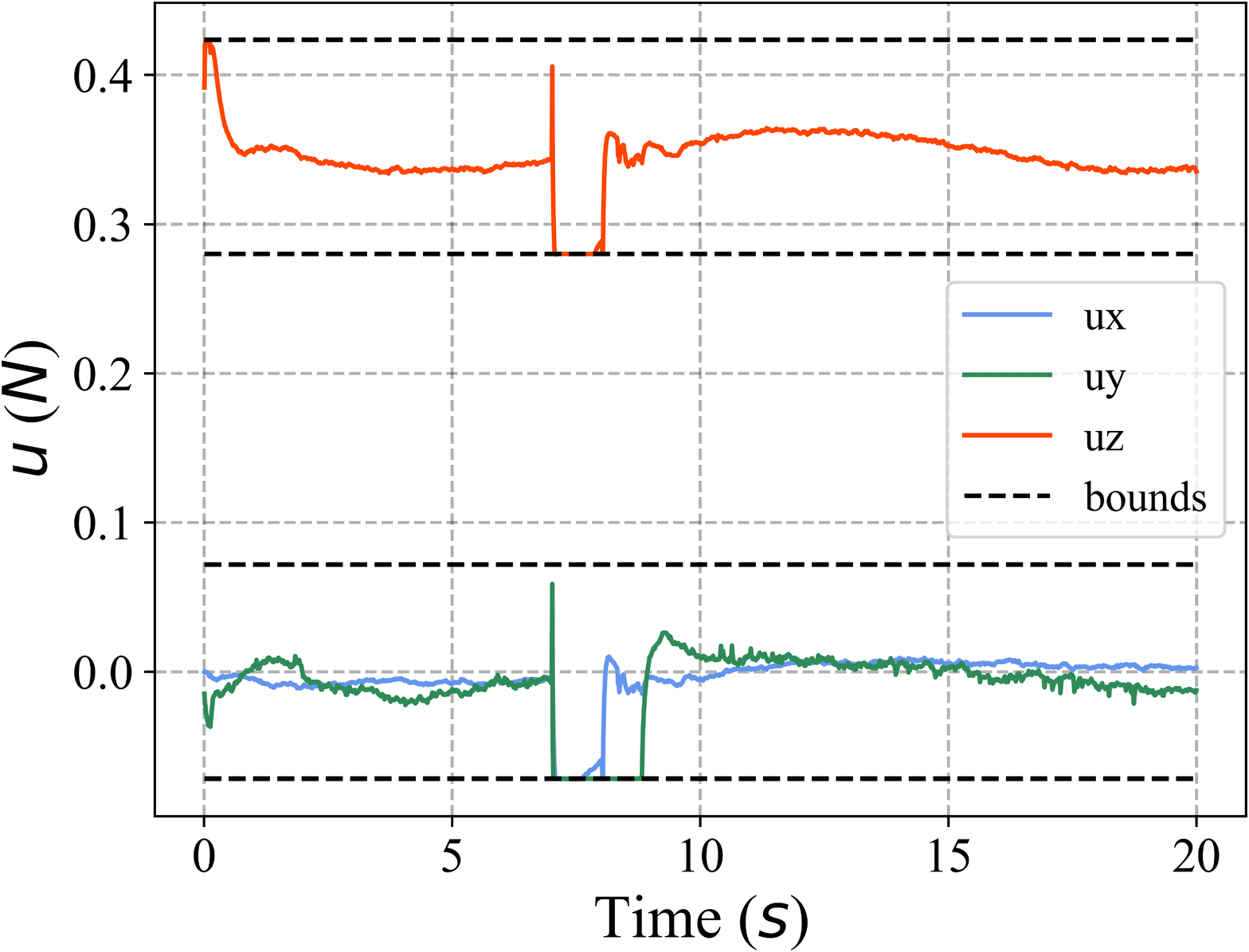}}
		\vspace{-3mm}
		\caption{(a) The evolution of parameter $\theta$ in the aggressive lemniscate path. (b) Control constraints are satisfied when following the lemniscate path with the proposed method.}	
		\vspace{-4mm}
	\end{figure}	
	\begin{table}[tp]
		\renewcommand{\arraystretch}{1.1}
		\scriptsize
		\caption{The RMSEs (in meter) with different levels of measurement noises and noise settings $\alpha$ in GPs.}	
		\label{table:GP_noise}
		\centering
		\vspace{-3mm}
		\begin{tabular}{ c | c c c c  }
			\hline
			Measurement noise & $\alpha = 0.0005$ &  
			$0.005$ & $0.05$  &  $0.5$\\
			\hline
			0.005 & \textbf{0.00852} &  \textbf{0.01346} & \textbf{0.01346} & \textbf{0.01345} \\
			0.05 & {0.01226} &  \textbf{0.01362} & \textbf{0.01498}  &\textbf{0.01518}\\
			0.1 & 0.01645 &  {0.01423}  & \textbf{0.01500}  & \textbf{0.01518}\\
			0.5 & 0.02847 &  0.28544  & 0.38680  &\textbf{0.01518}\\
			\hline
		\end{tabular} 
	\vspace{-4mm}
	\end{table}
	\begin{figure}[tp]
	 	\centering 
	 	\includegraphics[scale=0.05]{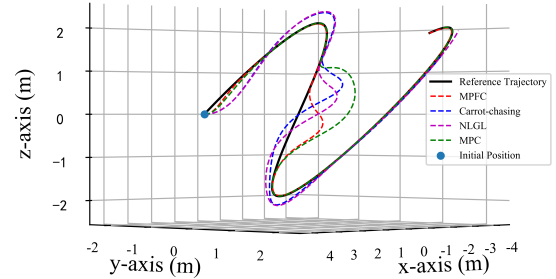}	
	 	\vspace{-3mm}
	 	\caption{Trajectories of the quadrotor following the aggressive lemniscate path. The proposed control scheme with MPFC as the high-level controller can achieve lower following error due to the predictive control of the system and reference evolution. Best viewed in color.}	
	 	\label{fig:8-trajs}
	 	\vspace{-4mm}
	\end{figure}	
	 
	A set of five paths, i.e., lemniscate, parabola, circle, cylindrical helix (CH) and conical spiral (CS), are used to test the path following performance, as shown in Fig. \ref{fig:all_paths}. These paths are parameterized by $\theta \in [0,20]$ with nominal velocity profiles by setting $\theta_{vel}(t)=1s^{-1}$. We start the reference target and initial position of the quadrotor on the path at $P(\theta=0)$. To quantify the following performance, we test for $T=20s$ and compare the root mean square error (RMSE) of the minimum distance from the quadrotor position to the path: 
	\vspace{-1mm}
	\begin{equation}
		\footnotesize
		\mathcal{E}=\sqrt{\frac{1}{M}\sum_{k=1}^{M}\|d_{min}\|^2},
		\vspace{-1mm}
	\end{equation}
	where the minimum distance is calculated as $d_{min}=\min_{\theta}{(x_1(t_k)-P(\theta))}$, and $M=T/dt=2000$ represents the total control steps in the discrete control.

	To evaluate the algorithm performance under unknown disturbances, a wind model in \cite{cole2018reactive} is utilized, consisting of a constant component $v_c$, a turbulent component $v_t$ and a wind gust component $v_g$, i.e., wind velocity $v_w=v_c+v_t+v_g$. The horizontal constant wind $v_c$ is randomly set from $3 m/s$ to $10 m/s$. The turbulence wind uses the von Kármán velocity model defined analytically in the specification MIL-F-8785C \cite{moorhouse1980us}, with the low-altitude model in the specification for the model parameters. The turbulent wind component $v_t$ can be generated following \cite{shinozuka1972monte} and \cite{deodatis1996simulation} with the model. The gusting profile $v_g$ is defined as a $1-cos$ model in the specification. We randomly generate $10$ sets of parameters of the wind model to evaluate the algorithm performance. 
	\vspace{-2mm}
	\subsection{Results}
	\vspace{-1.5mm}
	\subsubsection{Adaptability}
	In this subsection, we verify that $(i)$ the actual nonlinear system remains close to the ideal integrator model, and $(ii)$ the adaptive performance benefits from the designed LB-FBLC.
	
	White noises are added to the acceleration measurements to simulate the real sensor condition. The noise level on the measurements is considered in the GPs (5)-(7), corresponding to the noise level parameter $\alpha$ in the scikit-learn package of GPs. Table \ref{table:GP_noise} shows the following errors for the lemniscale path. When the parameter $\alpha$ in the GPs are properly chosen, the proposed method is robust to the measurement noises unless the measurement noises increase beyond a certain range. In the following numerical validation, white noise $w_i\sim N(0,0.005)$ and $\alpha = 5e$-$4$ are set.

	To validate $(i)$, we compare the actual trajectory of the quadrotor following the lemniscale path with the reference trajectory generated by MPFC, as shown in Fig. \ref{fig:linear_integrator}. The reference states satisfy an ideal integrator (\ref{eq:desired_integrator}). The actual system under unknown wind disturbances does behave close to the ideal integrator using the LB-FBLC. Fig. \ref{fig:GP} shows the estimations of wind disturbances $f_a$ on quadrotor using GPs. It can be seen that the actual disturbances lie in the uncertainty bound of the estimations. 
	
	We keep the high-level MPFC the same to fairly compare the designed low-level LB-FBLC with five controllers: a) a nominal feedback linearization controller (FBLC) with no GPs; b) a nominal feedforward linearization controller (FFLC) with no GPs; c) a learning-based feedforward linearization controller using GPs (LB-FFLC), which removes the PD feedback control from the LB-FBLC; d) the robust control method proposed in~\cite{helwa2019provably} (ROBUST). For the ROBUST method, we selected the parameter $\epsilon=0.1$. The Lyapunov function and PD control gain matrix in the ROBUST method are set the same as those in our method. The settings on GPs among the learning-based methods are the same for fairness of comparison. 
	
	The path following RMSEs shown in Table.\ref{table:adaptive_table} are averaged over five paths with or without disturbances. The proposed low-level LB-FBLC achieves the minimum RMSE in all cases. Both the LB-FBLC and LB-FFLC use GPs to learn the uncertain disturbances and achieve lower tracking error than the nominal FBLC and FFLC methods under disturbances. It shows that the proposed approach benefits from using GPs to compensate the unknown disturbances. Besides, the LB-FBLC/FBLC achieves a smaller RMSE compared with the LB-FFLC/FFLC. It demonstrates that the feedback PD control design in the proposed approach is effective. Table.\ref{table:adaptive_table} also shows that the proposed approach can achieve lower RMSE compared to the ROBUST method. The results show that the designed adaptive control law, with the form and the condition of stability different from the ROBUST method, can effectively reduce the path following error under the unknown wind disturbances.

	\subsubsection{Predictivity}
	The benefits of the predictive control inherent in the designed MPFC are demonstrated in this part. For a fair comparison, we keep the LB-FBLC as the low-level controller and compare the high-level MPFC against two baseline path following control algorithms: a) Carrot-chasing, b) Nonlinear Guidance Law (NLGL), and a trajectory control algorithm: c) MPC. In the Carrot-chasing method, the reference target is chosen at a constant distance $D_1$ ahead from the path point which is computed by projecting the quadrotor position to the path. In the NLGL, the reference target is calculated as the point on the path, which is at a distance $D_2$ from the quadrotor. The $D_1$ and $D_2$ are set for the best following performance. Since the velocity profile of the path is predefined with the evolution velocity of the parameter $\theta_{vel}(t)=1 s^{-1}$, we can compare the trajectory control performance with the predefined time-parameterized reference trajectory $\mathcal{P}(\theta(t))=\mathcal{P}(t)$, to illustrate the advantage of path following described in Section~\ref{sec:introd}. The MPC uses the same double integrator predictive model and parameter settings as in the MPFC.
	 
	As shown in Table.\ref{table:adaptive_table}, the proposed method with MPFC	as the high-level controller can largely reduce the averaged RMSE compared with the Carrot-chasing and NLGL method in all cases. To assess the robustness to irresistible disturbances, a large wind gust of around $20 m/s$ is added to the wind $v_w$ from $7s$ to $8s$ to push the quadrotor away from the path. The proposed approach reduces both the maximum following error and the RMSE under disturbances, as shown in table \ref{table:table_predictivity} and Fig.\ref{fig:trajs}, respectively, compared with the Carrot-chasing and the NLGL method. Without disturbances, the following errors can also be reduced largely with the MPFC as the high-level controller. To obtain an intuitive view on the control performance under disturbances, Fig. \ref{fig:8-trajs} shows the trajectories of the quadrotor following the aggressive lemniscate path using different high-level controllers, with the position errors shown in Fig. \ref{fig:disturbance_error}. It can be seen that the quadrotor is blown away but soon converges back to the path rapidly with smaller deviation and obviously less overshooting using the proposed method compared with the baseline NLGL and Carrot-chasing methods. Lower following errors can also be observed in the sharp turns.

	Figure \ref{fig:theta_v} shows the evolution velocity of the reference parameter $\theta$. The MPC tracks the time-parameterized trajectory with higher tracking errors after the quadrotor deviates from the path as shown in Fig. \ref{fig:mpc}. With MPFC as the high-level controller, the evolution of the reference target slows down when the quadrotor deviates from the path at around 7$s$ and sharp turns. It reveals that the path following error can be largely reduced with the help of optimization of reference target evolution in the MPFC. Figure \ref{fig:control_Inputs} illustrates the control constraints of $u$ can be satisfied with the designed learning-based control scheme. 
	 
	\vspace{-2mm}
	\section{Conclusions}
	\label{sec:con}
	\vspace{-1mm}
	
	In this paper, a novel learning-based predictive control scheme is presented for nonlinear systems to accurately follow paths under uncertain environmental disturbances. A low-level LB-FBLC with GPs is designed to track the reference states accurately under disturbances with a probabilistic stability guarantee. A high-level MPFC exploits an integrator system model and a virtual linear path dynamics model to simultaneously optimize the reference target revolution, and provides the reference states and controls for the LB-FBLC. Simulation results show that the proposed control scheme can successfully drive a quadrotor to accurately follow various geometric paths under different unknown wind disturbances. Both the maximum and the mean following errors are shown to be effectively reduced using the proposed method. The quadrotor with the proposed control scheme exhibits predictive ability when following aggressive paths and robustness to the wind disturbances. In future work, estimation delay and limited update frequency will be considered, as well as hardware experiments under real conditions.
	\vspace{-2mm}

	\bibliographystyle{IEEEtran}
	\bibliography{egbib}
	
\end{document}